# Evaluación de algoritmos bioinspirados para la solución del problema de planificación de trabajos [1]

## Evaluation of bioinspired algorithms for the solution of the job scheduling problem.

Edson Flórez[2], Nelson Díaz[3], Wilfredo Gómez[4], Lola Bautista[5], Darío Delgado[6]

Universidad Industrial de Santander, Bucaramanga, Colombia.

---

[2] Ingeniero de Sistemas, Universidad Industrial de Santander UIS. Magister en Ingeniería de Sistemas e Informática, Universidad Industrial de Santander UIS. Estudiante de Doctorado en Informática, Laboratoire d'Informatique, Signaux et Systèmes de Sophia-Antipolis (I3S), Université Nice Sophia Antipolis 2000, route des Lucioles. Les Algorithmes, bât. Euclide B. 06900, Sophia Antipolis, France. ORCID: N.D. Correo institucional: florez@i3s.unice.fr, edson.florez@correo.uis.edu.co.

[3] Ingeniero de Sistemas, Magister en Ingeniería de Sistemas e Informática, Estudiante de Doctorado en Ingeniería (Ing.Eléctrica, Electrónica y Gestión & Desarrollo) Universidad Industrial de Santander UIS. Bucaramanga, Colombia. Cra 27 Calle 9 Ciudad Universitaria. PBX: 6344000. ORCID: N.D. Correo institucional: epsilon1530@gmail.com

[4] Ingeniero de Sistemas e Informática, Universidad Industrial de Santander UIS. Magister en Ingeniería de Sistemas e Informática, Universidad Industrial de Santander UIS. Estudiante de Doctorado en Administración de Empresas, Universidad Politécnica de Valencia. Asesor técnico en innovación de la Iniciativa Apps.Co, Ministerio de Tecnologias de la Información y las Comunicaciones MINTIC, Cra. 8 #Entre Calles 12 y 13, Bogotá. ORCID: http://orcid.org/0000-0002-9332-7760 Correo institucional: wgomezb@mintic.gov.co.

[5] Ingeniero de Sistemas, Universidad Industrial de Santander UIS. Master of Science in Computer Engineering (Electrical and Computer Engineering), Universidad de Puerto Rico: Mayagüez, Puerto Rico. Doctor en Automatique et Traitement de Signaux et des Images (Ecole Doctorale des Sciences et Technologies de l'Information et de la Communication) Université de Nice Sophia Antipolis: Nice, Provence-Alpes-Côte d'Azu, France. Docente planta de la Escuela de Ingeniería de Sistemas e Informática en la Universidad Industrial de Santander: Bucaramanga, Colombia. Cra 27 Calle 9 Ciudad Universitaria. PBX: 6344000. ORCID: orcid.org/0000-0002-3853-007X Correo institucional: lxbautis@uis.edu.co

[6] Ingeniero de Sistemas, Magister en Ingeniería de Sistemas e Informática, Doctor en Ingeniería (Ing.Eléctrica, Electrónica y Gestión & Desarrollo) Universidad Industrial de Santander UIS. Investigador en la Universidad Nacional Abierta y a Distancia – UNAD, Calle 14S #14 23 Restrepo Bogotá PBX: (1) 3443700. ORCID: N.D. Correo institucional: dario.delgado@unad.edu.co

Edson Flórez, Nelson Díaz, Leydy Luna, Wilfredo Gómez, Lola Bautista
Evaluación de algoritmos bioinspirados para la solución del problema de planificación de trabajos

**Resumen**

En el presente trabajo se utilizaron Metaheurísticas de inspiración biológica, como sistemas inmunes artificiales y los algoritmos de colonias de hormigas que se basan en características y comportamientos de los seres vivos aplicables en el área computacional. Se presenta una evaluación de soluciones bioinspiradas para el problema de optimización combinatoria de planificación de trabajos, denominado Job shop Scheduling, cuyo objetivo es encontrar una configuración o secuencia de trabajos que requiera la menor cantidad de tiempo para ser ejecutada en las máquinas disponibles. El desempeño de los algoritmos fue caracterizado y evaluado para instancias de referencia del problema de Job Shop Scheduling, comparando la calidad de las soluciones obtenidas respecto a la mejor solución conocida (BKS por sus siglas en inglés) de los métodos más eficaces. Las soluciones fueron valoradas en dos aspectos, en calidad tomando como referente el makespan, que corresponde al tiempo que tardan en realizarse todos los trabajos, y como métrica de desempeño se tomó el número de evaluaciones que realiza el algoritmo para obtener la mejor solución.

*Palabras clave*: algoritmos bioinspirados, job shop scheduling, metaheurísticas, optimización combinatoria, planificación de tareas.

**Abstract.**

In this research we used bio-inspired metaheuristics, as artificial immune systems and ant colony algorithms that are based on a number of characteristics and behaviors of living things that are interesting in the computer science area. This paper presents an evaluation of bio-inspired solutions to combinatorial optimization problem, called the Job Shop Scheduling or planning work, in a simple way the objective is to find a configuration or job stream that has the least amount of time to be executed in machine settings. The performance of the algorithms was characterized and evaluated for reference instances of the job shop scheduling problem, comparing the quality of the solutions obtained with respect to the best known solution of the most effective methods. The solutions were evaluated in two aspects, first in relation of quality of solutions, taking as reference the makespan and secondly in relation of performance, taking the number evaluations performed by the algorithm to obtain the best solution.

**Introducción**

Los problemas de secuenciación de tareas aparecen constantemente en la vida real en numerosos ambientes productivos y de servicios. Son problemas en los cuales se requiere organizar la ejecución de trabajos que compiten entre sí por el uso de un conjunto finito de recursos, y que a su vez están sujetos a un conjunto de restricciones.

Edson Flórez, Nelson Díaz, Leydy Luna, Wilfredo Gómez, Lola Bautista
Evaluación de algoritmos bioinspirados para la solución del problema de planificación de trabajos

Estos problemas son combinatorios y se caracterizan por pertenecer a la clase NP-Duro (M. R. Garey and D. S. Johnson,1979), en otras palabras, problemas en los que el tiempo de cómputo necesario para resolverlos crece de forma exponencial conforme aumenta el tamaño del problema. Por esta razón los investigadores han desarrollado distintos métodos o algoritmos que procuran soluciones efectivas, ya sea de forma determinística o no determinística (C. A. C. Coello, D. R. Cortés, and N. C. Cortés,2004).

El objetivo de la optimización es maximizar o minimizar los criterios sujetos a las restricciones en donde cada trabajo es una secuencia de operaciones, cada una con una máquina y tiempo de procesamiento determinados. Las soluciones factibles deben cumplir con las restricciones a las que está sujeto el problema de Job Shop Sheduling, como es la de respetar la precedencia entre operaciones que determina la secuencia tecnológica sin interrumpir ninguna operación hasta su finalización (A. Manne,1960).

Debido a su complejidad, los problemas de programación de tareas precisan de algoritmos de búsqueda eficientes para encontrar soluciones aceptables en tiempos razonables (M. R. Garey, D. S. Johnson, and R. Sethi,1976). En la literatura se pueden encontrar aproximaciones a los problemas de secuenciación de tareas basadas en los algoritmos de búsqueda heurística propios de áreas como la Investigación Operativa y la Inteligencia Artificial (F. Ghedjati,2010), (C. Blum and A. Roli,2003).

Para la solución de este problema se han usado diferentes algoritmos que incluyen técnicas como: procesos de vuelta atrás (backtracking), ramificación y poda (branch and bound), programación dinámica [7], las heurísticas de construcción voraz o heurísticas greedy (M. G. C. Resende and C. C. Ribeiro,2010); adicionalmente en los últimos años se han usado metaheurísticas como son: Enfriamiento simulado, búsqueda tabú (D. De Inform, T. Universidad, and S. Mar,2006), búsqueda local iterativa (*Iterated Local Search*), algoritmos de búsqueda local con vecindario variable (*Variable Neighborhood Search*), GRASP (*Greedy Randomized Adaptative Search Procedures*) (M.Tupia,2005), (S. Binato, W. J. Hery, and D. M. Loewenstern,2000) y finalmente los algoritmos bioinspirados, entre los que destacamos las técnicas utilizadas en este artículo como lo son los algoritmos de colonia de hormigas (M. Ventresca and B. M. Ombuki,2004) y algoritmos inmunes (Z. Problemler, Y. Ba, and N. D. E. Yen,2007).

En este desarrollo se destaca la caracterización de dos técnicas bioinspiradas, una probabilística y la otra evolutiva, lo que permite establecer un escenario común de comparación utilizando representaciones de las soluciones e instancias de evaluaciones que faciliten dicho contraste. Adicional se presenta una referencia actualizada para la configuración de los parámetros iniciales de los algoritmos en problemas de índole combinatoria y finalmente se detalla la experiencia de implementación de este tipo de técnicas bioinspiradas con la inclusión y modificación de operadores y la hibridación de metaheurística como en el caso de algoritmos de optimización de colonia de hormigas con búsqueda tabú y algoritmos voraces. Este desarrollo pretende enriquecer el panorama de documentación metodológica para la implementación de metaheurística y de esta manera aportar y motivar al desarrollo de nuevos modelos bioinspirados en los que se pueda mejorar el desempeño de este tipo de técnicas.

Este artículo se encuentra organizado de la siguiente forma: se inicia con una presentación de trabajos relacionados con Job Shop Scheduling (JSP) en la siguiente sección se realiza una descripción formal del problema Job Shop Scheduling. El artículo continúa con una descripción de las instancias del problema empleados para el presente estudio. En la siguiente sección se describe las técnicas desarrolladas como el algoritmo inmune y el algoritmo de colonia de hormigas, posteriormente se presentan y discuten los resultados. Este trabajo finaliza con las conclusiones generadas en el proceso de desarrollo e investigación.

### 1. JOB SHOP SCHEDULING

Job Shop Scheduling Problem (JSP) es un problema de secuenciación de tareas, para el cual un conjunto finito de trabajos es procesado sobre un conjunto finito de máquinas. Cada trabajo se caracteriza por un orden fijo de las operaciones, cada una de las cuales será procesada en una máquina específica durante un tiempo determinado. Cada máquina puede procesar a lo más un trabajo en un tiempo y una vez que un trabajo ha iniciado sobre una máquina se debe completar su procesamiento sobre esa máquina por un tiempo ininterrumpido. Un Calendario es una asignación de operaciones en intervalos de tiempo sobre las máquinas. El objetivo de JSP es encontrar un calendario que minimice la función objetivo. Para el presente caso de estudio la función objetivo será el makespan, que es el tiempo en completar todos los trabajos, es decir, la longitud del calendario desde que empieza a ejecutarse el primer trabajo hasta que finaliza el último trabajo.

Formalmente, JSP puede ser definido como se muestra en el trabajo de Witkowski et al(T. Witkowski, P. Antczak, and A. Antczak,2010). Dado un conjunto $M$ de máquinas ($|M|$ denota el tamaño de $M$) y un conjunto $J$ de trabajos ($|J|$ denota el tamaño de $J$), sean $\sigma_1^j \prec \sigma_2^j \prec \cdots \prec \sigma_{|M|}^j$ el orden de un conjunto de $|M|$ operaciones del trabajo $j$, donde $\sigma_k^j \prec \sigma_{k+1}^j$ indica que la operación $\sigma_{k+1}^j$ solo puede empezar el procesamiento después de completar la operación $\sigma_k^j$. Sea $O$ el conjunto de operaciones. Cada operación es definida por dos parámetros: $M_k^j$ es la maquina sobre la cual $\sigma_k^j$ es procesada y $p_k^j = p(\sigma_k^j)$ es el tiempo de procesamiento de la operación $\sigma_k^j$. Se denota como $t(\sigma_k^j)$ el tiempo de inicio de la $k-esima$ operación $\sigma_k^j \in O$. Una formulación de programación disyuntiva para el JSP se muestra a continuación:

min $C_{max}$ sujeto a:

$C_{max} \geq t(\sigma_k^j) + p(\sigma_k^j)$, para toda $\sigma_k^j \in O$,

(1a). $t(\sigma_k^j) \geq t(\sigma_l^j) + p(\sigma_l^j)$, para toda $\sigma_l^j \prec \sigma_k^j$,

(1b). $t(\sigma_k^j) \geq t(\sigma_l^i) + p(\sigma_l^i) \vee$

$t(\sigma_l^i) \geq t(\sigma_k^j) + p(\sigma_k^j)$, para todo $i,j \in J \ni M_{\sigma_l^i} = M_{\sigma_k^i}$,

$t(\sigma_k^j) \geq 0$, para toda $\sigma_k^j \in O$

Donde $C_{max}$ es el makespan a ser minimizado.

Una solución factible puede ser construida de una permutación de los trabajos del conjunto $J$ sobre cada una de las máquinas del conjunto $M$, observando las restricciones de precedencia, la restricción de que cada máquina puede procesar solo una operación a la vez y la restricción que garantiza el procesamiento de una operación de manera ininterrumpida en una máquina hasta ser completada. Cada conjunto de permutaciones tiene un correspondiente calendario. Por tanto, el objetivo del JSP es encontrar un conjunto de permutaciones con el makespan mínimo. El presente trabajo aborda JSP mono-objetivo dado que solo minimiza el makespan.

Para probar la implementación se usó la familia de instancias del problema de Job Shop Scheduling conocidas como instancias de Lawrence (LA) (S. R. Lawrence,1984). Consta de 40 problemas de 8 diferentes tamaños propuestos (M. Ventresca and B. M. Ombuki,2004): 10 x 5, 15 x 5, 20 x 5, 10 x 10, 15 x 10, 20 x 10, 30 x 10 y 15 x 15.LA (D. Applegate y W. Cook,1991) es una de las familias más comúnmente utilizadas para probar el desempeño de JSP. Cada instancia se compone de una fila de descripción y varias filas con valores enteros. Cada fila de valores enteros corresponde a un trabajo. El trabajo es un conjunto de parejas conocido como operaciones, la pareja es integrada por el número de la máquina y tiempo de procesamiento en dicha máquina.

## 2. ALGORITMOS BIOINSPIRADOS

La computación bioinspirada se ha soportado en el creciente desarrollo tecnológico gracias a los grandes avances en la electrónica y hardware computacional, y ha venido avanzando sobre los métodos de computación clásicos, utilizando procedimientos que reproducen ciertas propiedades inspiradas en la biología con el fin de diversificar los resultados obtenidos en la medida que se obtengan mejores resultados. Esta disciplina, que está íntimamente vinculada al campo de la Inteligencia Artificial, engloba varios enfoques, tales como: los algoritmos evolutivos (AE) (A. E. Eiben and J. E. Smith,2003), la optimización de colonia de hormigas (ACO) (M. Dorigo and G. Di Caro,1999), la optimización de enjambre de partículas (PSO) (J. Kennedy and R. Eberhart,1995) entre otros. Particularmente en el presente trabajo se utilizaron 2 técnicas de diferentes características, primero un algoritmo de Colonia de Hormigas, que corresponde a una metaheurística probabilística que se basa en el comportamiento natural de las hormigas cuando estas se dirigen a buscar alimentos, y en segundo lugar un algoritmo inmune artificial que corresponde a un algoritmo evolutivo que se basa en la dinámica del sistema inmune de los vertebrados para detectar y eliminar posibles amenazas para el organismo.

## 2.1. ALGORITMO ELITISTA DE COLONIA DE HORMIGAS (EAS)

En los algoritmos de la familia Ant Colony Optimization, el comportamiento de las hormigas se simula con un agente virtual que tiene la capacidad de explorar un espacio de búsqueda limitado y obtener información acerca del entorno que lo rodea. La hormiga artificial (k) se mueve de un nodo a otro (del nodo origen i al nodo destino j), construyendo paso a paso una solución que se va guardando en la memoria Tabú (que almacena información sobre la secuencia de nodos o ruta seguida hasta el momento t), que termina cuando alcanza alguno de los estados de aceptación definidos por el objetivo del problema.

Así, en cada iteración, las hormigas pueden construir soluciones aproximadas a problemas complejos como los de secuenciación, asignación, planificación o programación:

1) $\eta_{ij}$ Información heurística que mide la preferencia heurística de moverse desde el nodo *i* hasta el nodo *j*, al recorrer la arista $a_{ij}$. Las hormigas no modifican esta información durante la ejecución del algoritmo.

2) $\tau_{ij}$ Información de los rastros de feromona artificiales, que mide la "deseabilidad aprendida" del movimiento de *i* a *j*. Esta información se modifica durante la ejecución del algoritmo dependiendo de las soluciones encontradas por las hormigas para reflejar la experiencia adquirida por estos agentes.

La versión de ACO desarrollada para el presente trabajo (E. Flórez, W. Gómez, and L. Bautista,2013) refuerza el rastro de feromona de la mejor ruta encontrada en cada iteración. A las aristas de la mejor solución generada por una de las hormigas, se tiene mayor probabilidad de depositar más feromona por medio de todas las otras hormigas.

En este algoritmo las hormigas artificiales realizan una construcción probabilística de soluciones en cada ciclo, para lo cual se requiere representar el problema por medio de un grafo, en el que las hormigas se mueven a lo largo de cada arista de un nodo a otro para construir caminos que representan soluciones, desde un nodo inicial seleccionado aleatoriamente. Las siguientes elecciones del próximo nodo en este camino se hacen de acuerdo a la regla de transición de estado:

$$p_{ij} = \frac{(\tau_{ij})^\alpha (\eta_{ij})^\beta}{\sum_{l \in Tabu_k} (\tau_{il})^\alpha (\eta_{il})^\beta} \quad (2)$$

$$si\ j,l\ \notin Tabu_k$$

Donde los parámetros $\alpha$ y β determinan la influencia de los valores de la información de la feromona ($\tau$) y de la información heurística (η) respectivamente, sobre la decisión de cada hormiga ($k$). Se procura que las aristas con gran cantidad de feromona sean las más visibles, teniendo una probabilidad de transición mayor a las aristas de los otros nodos del conjunto de operaciones realizables. Para tener un algoritmo equilibrado (con un apropiado ajuste), los parámetros $\alpha$ y $\beta$ deben tener valores adecuados, evitando valores cercanos a cero, porque si $\alpha = 0$, solo la información heurística indicaría que posibles elementos de la solución tendrán mayor probabilidad de ser seleccionados, lo que corresponde a un algoritmo greedy (voraz) estocástico, y si $\beta = 0$, solo será relevante la cantidad de feromona. En ambos casos las hormigas podrían estancarse en un óptimo local, generando la misma solución en cada iteración, sin oportunidades de encontrar una mejor solución que sea la solución óptima global. Estos parámetros están configurados normalmente en valores enteros entre 1 y 5 (E. Flórez, W. Gómez, and L. Bautista, 2013), pero en este caso los relacionaremos de la siguiente forma $\beta = (1 - \alpha)$ con $\alpha \in (0,1]$, como una distribución probabilista uniforme.

La cantidad de feromona $\tau_{ij}(t)$ presente en cada arista del camino en la generación $t$ está dada por la siguiente ecuación:

$$\tau_{ij}(t) = \sum_{k=1}^{n} \Delta \tau_{ij}^{k} + \rho * \tau_{ij}(t - 1) \quad (3)$$

Donde $\tau_{ij}^{k}(t)$ es la contribución de la hormiga $k$ a la feromona total en la generación $t$, y $\rho$ es la tasa de evaporación de la feromona. La razón para incluir la tasa de evaporación es que la feromona antigua no debería tener mucha influencia en las decisiones futuras de las hormigas.

La cantidad de feromona con la que contribuye cada hormiga depende de la calidad de la solución obtenida, siendo inversamente proporcional al costo de la función objetivo de la solución, así:

$$\Delta \tau_{ij}^{k} = \frac{Q}{L_k} \quad (4)$$

Donde $Q$ es una constante y $L_k$ es la longitud del makespan de la solución obtenida por la hormiga $k$.

Hasta el punto anterior las ecuaciones son idénticas a las del Ant System, la modificación planteada por el mismo Dorigo (M. Dorigo and G. Di Caro,1999) permite acelerar la convergencia del algoritmo, aumentando la visibilidad del rastro de feromona en todas las aristas del camino más corto, pasando con todas las hormigas elitistas ($e$) del sistema (E. Flórez, W. Gómez, and L. Bautista,2013) (Ver ecuación 5), que son un número determinado de hormigas destinadas solo a fortalecer el rastro de feromona en los mejores caminos. Por lo tanto, la ecuación 3 para el mejor camino construido en cada ciclo es reemplazada por:

$$\Delta \tau_{ij}^{k} = \frac{Q}{L_k} * e \quad (5)$$

### 3.2. ALGORITMO INMUNE ARTIFICIAL DE SELECCIÓN CLONAL (CLONALG)

Castro y Von Zuben (L. N. De Castro, C. Brazil, and F. J. Von Zuben,2000) desarrollaron un algoritmo inspirado en Sistemas Inmunes Artificiales basados en la abstracción de selección clonal, el cual ha sido utilizado para realizar tareas de emparejamiento de patrones (W. A. Gómez Bueno, M. I. Cuadrado Morad, and H. Arguello Fuentes,2012)(U. Garain, M. P. Chakraborty, and D. Dasgupta,2006) y optimización de funciones multimodal (N. C. Cortes and C. A. C. Coello,2003)(F. O. de França, F. J. Von Zuben, and L. N. de Castro,2005), entre otras aplicaciones (L. N. de Castro and F. J. Von Zuben,2002)(L. N. De Castro, C. Brazil, and F. J. Von Zuben,2000). Este algoritmo se conoce como Clonalg (J. Brownlee,2007) y la implementación computacional de este principio presenta la siguiente estructura general(N. E. D. DIAZ and L. J. L. Martínez,2012):

1) Generar aleatoriamente un conjunto (P= $Ab_i\ i = 1, ..., n$) donde $Ab_i$ es el conjunto de soluciones potenciales, dicho conjunto lo componen, haciendo la analogía a términos biológicos, la representación de los anticuerpos de las células de memoria M y la representación de los anticuerpos producidos por los demás linfocitos del organismo.

En el algoritmo propuesto para la solución del problema Job Shop Scheduling (N. Diaz, J. Luna, W. Gomez, and L. Bautista,2013), cada anticuerpo corresponde a una sarta de valores enteros que representa a un calendario candidato a ser solución de la instancia que se está evaluando durante dicha ejecución.

Cada uno de los anticuerpos o calendarios candidatos se crean de manera aleatoria, y se garantiza su factibilidad mediante la verificación de las restricciones del problema durante el proceso de generación.

2) Determinar los *n* mejores individuos de la población de anticuerpos, midiendo el grado de afinidad de cada uno con la representación del antígeno (evaluando cada solución con respecto a la función a optimizar).

Para el algoritmo propuesto la afinidad $f$ de anticuerpos está dada por la siguiente ecuación:

$$f = 1 - \frac{\text{makespan}}{\text{rango}} \qquad (6)$$

Donde el rango está dado por:

$$\text{rango} = \max(\text{makespan}) - \min(\text{makespan}) \quad (7)$$

3) Reproduzca los *n* mejores individuos (dependiendo qué tan bien cada solución se comporta en relación con el valor de la función), generando una población temporal de clones (C).

Para el algoritmo propuesto, la cantidad $n$ de clones para cada anticuerpo corresponde a un parámetro del algoritmo, el cual está dado por la ecuación:

$$N_c = (\beta * N) \qquad (8)$$

Donde $\beta$ es un factor de clonación y $N$ es la cantidad de anticuerpos.

Para el algoritmo desarrollado se tomó un rango entre 0 y 1 para realizar las pruebas del análisis de sensibilidad en lo referente al factor de clonación; después de dicho análisis se determinó como factor de clonación el valor de 0,1.

4) Someta a la población de clones a un proceso de hipermutación, haciendo que la variación sea inversamente proporcional al grado de afinidad del anticuerpo; este proceso da lugar a una nueva población de clones (C*).

En el algoritmo propuesto los clones generados sufren un proceso de hipermutación la cual es inversamente proporcional a la afinidad del antígeno (entre mayor afinidad, menor es la tasa de mutación), dicha probabilidad de mutación está dada por:

$$p = e^{(-\rho * f)} \qquad (9)$$

Donde, $\rho$ es el factor de mutación y $f$ la afinidad.

Al fijar este parámetro se confirmó la noción planteada por Cortés (N. C. Cortes and C. A. C. Coello,2003) donde *"El impacto que tiene la mutación en un anticuerpo para generar un nuevo individuo es mínima, ya que se aplica de tal forma que sólo se realiza un cambio en la cadena"*.

5) Seleccione los individuos mejorados del conjunto C* para estructurar un conjunto de "memoria".

6) Reemplace con los mejores individuos del conjunto C* a los *d* anticuerpos con menos afinidad de la población inicial.

Adicionalmente se cuenta con el parámetro denominado número randómico de células, que corresponde al porcentaje de individuos respecto a la población que serían generados aleatoriamente con el fin de mantener la diversidad en la población. El rango se encuentra entre 0 y 100% respecto a la población. El valor calculado que optimiza el makespan para la mayoría de instancias fue del 10%. Este valor cobra sentido pues valores pequeños en este parámetro no alteran negativamente la diversidad en la población, lo cual haría que el algoritmo se estancara en una región del espacio de búsqueda.

## 3. METODOLOGÍA

Para la realización del presente trabajo se plantearon las siguientes etapas:

1) Caracterización del Problema: En esta etapa se estudió la función objetivo del problema a resolver, se analizaron sus variables y la dinámica de comportamiento de las mismas. Con esto se pudo identificar cómo cada algoritmo, teniendo en cuenta sus características podría establecer soluciones parciales que pudiesen ser mejoradas en cada uno de los ciclos, recorridos o iteraciones. Para esto se revisó la literatura de metaheurísticas, y se establecieron los referentes de soluciones de problemas combinatorios y específicamente problemas de planificación.

2) Representación de la solución: Esta es una de las etapas fundamentales en el desarrollo, porque permite establecer cómo la inspiración biológica se puede caracterizar en las variables de los modelos matemáticos de los problemas de optimización. A partir de ahí en el caso del algoritmo inmune se logró la representación de los anticuerpos como sartas con valores reales que corresponden a la configuración de un calendario que es una solución candidata. Para el caso del algoritmo de hormigas, las hormigas construyen configuraciones posibles de calendarios como las rutas que se guardan en una memoria tabú, eso se codifica en sartas que se evalúan de acuerdo a las restricciones del problema.

3) Determinación de las instancias: Para la evaluación de este tipo de desarrollos se cuentan con familias de instancias creadas artificialmente para valorar el desempeño de algoritmos o metaheurísticas de optimización, la clave de esta etapa fue seleccionar una familia que brindara variedad, complejidad y buena difusión en la literatura para poder realizar contrastes con otros trabajos. De ahí que se valoraron 5 tipos distintos de familias y se seleccionaron las instancias de Lawrence, que son muy comunes para evaluar los algoritmos en este problema.

4) Implementación de los Algoritmos: Se analizaron las diferentes abstracciones de los algoritmos seleccionados y se determinaron como base dos técnicas, en el caso de los algoritmos inmunes artificiales se trabajó con el algoritmo de selección clonal al que se le realizaron las adaptaciones correspondientes en los operadores para que funcionara adecuadamente en el problema seleccionado y con la codificación especificada. Para el caso de la metaheurística de colonia de hormigas se tomó como base el algoritmo de optimización por colonia de hormigas con algunas modificaciones como la inclusión del operador elitista y el apoyo de una búsqueda tabú que terminaron haciendo más eficiente al algoritmo genérico de optimización por colonia de hormigas.

5) Configuración de los algoritmos: Con un análisis de sensibilidad se establecieron los valores de los parámetros de entrada para los dos algoritmos (EAS, CLONALG), dichos valores se presentan en la Tabla 1 a continuación:

**TABLA 1.** Parámetros iniciales algoritmo EAS

| Parámetro EAS | Valor |
| --- | --- |
| Influencia de la Feromona ($\alpha$) | 0,2 |
| Influencia de la heuristic information ($\beta$) | 0,8 |
| Evaporación de la Feromona ($\rho$) | 0,7 |
| Feromona Inicial ($\tau_0$) | 0,002 |
| Feromona Ganada (Q) | 0,001 |
| Número de Ciclos | 1000 |

**TABLA 2.** Parámetros iniciales algoritmo CLONAG

| Parámetro CLONALG | Valor |
| --- | --- |
| Población anticuerpos | 100 |
| Numero de generaciones | 300 |
| Factor de mutación | 0,1 |
| Factor de clonación | 0,1 |
| Factor de generación | 10% |

6) Evaluación: Se establecieron dos criterios de evaluación, calidad y eficiencia de la solución. Para medir el desempeño de los algoritmos se realizaron comparaciones del número de evaluaciones de la función objetivo, que en el caso del problema corresponde al cálculo del makespan. Para medir la calidad de las soluciones, se realizó una comparación directa del makespan obtenido por el algoritmo propuesto en cada una de las instancias en 50 ejecuciones y se comparó con el mejor conocido (BKS) (Ver Tabla 3). Las medidas estadísticas que se reportan son el promedio $\bar{x}$, la varianza $S^2$ y la desviación estándar. De igual manera se valoró la cantidad de ciclos o iteraciones que necesitó cada algoritmo para llegar a la mejor solución, con esto se estableció una métrica de eficiencia, para estas se aplicaron de igual manera las medidas estadísticas mencionadas anteriormente.

### 4. PRUEBAS Y RESULTADOS

Las técnicas fueron programadas en lenguaje Java y las pruebas se desarrollaron en un equipo portátil con sistema operativo Windows 7, procesador Intel Core i5 primera generación a 2.53 Ghz y 4Gb de memoria RAM. Estas prestaciones fueron suficientes y no se requirió de apoyo adicional en procesamiento ni memoria.

En la Tabla 3 y en la Figura 1, se observa el compendio de los resultados en cada instancia de Lawrence, se listan el mejor makespan $C_{max}$ obtenido por cada algoritmo, el error relativo frente al makespan de la mejor solución conocida (*Best Known Solution*, BKS) en la literatura (D. Cortés Rivera, 2004), además se incluye la mejor solución conocida y el tamaño de cada instancia.

**TABLA 3.** Resumen de errores relativos de los algoritmos para cada instancia del JSP (*Best Known Solution* [BKS], Algoritmo Colonia de Hormigas Elitista [EAS], Algoritmo de Selección Clonal [CLONALG])

| Instancia | Tamaño | BKS | EAS | | CLONALG | |
|---|---|---|---|---|---|---|
| | | | $C_{max}$ | Error relativo % | $C_{max}$ | Error relativo % |
| LA01 | 10 x 5 | **666** | **666** | 0,00 | **666** | 0,00 |
| LA02 | | **655** | 669 | 2,14 | **655** | 0,00 |
| LA03 | | **597** | 617 | 3,35 | 603 | 1,01 |
| LA04 | | **590** | 595 | 0,85 | **590** | 0,00 |
| LA05 | | **593** | **593** | 0,00 | **593** | 0,00 |
| LA06 | 15 x 5 | **926** | **926** | 0,00 | **926** | 0,00 |
| LA07 | | **890** | **890** | 0,00 | **890** | 0,00 |
| LA08 | | **863** | **863** | 0,00 | **863** | 0,00 |
| LA09 | | **951** | **951** | 0,00 | **951** | 0,00 |
| LA10 | | **958** | **958** | 0,00 | **958** | 0,00 |
| LA11 | 20 x 5 | **1222** | **1222** | 0,00 | **1222** | 0,00 |
| LA12 | | **1039** | **1039** | 0,00 | **1039** | 0,00 |
| LA13 | | **1150** | **1150** | 0,00 | **1150** | 0,00 |
| LA14 | | **1292** | **1292** | 0,00 | **1292** | 0,00 |
| LA15 | | **1207** | 1212 | 0,41 | **1207** | 0,00 |
| LA16 | 10 x 10 | **945** | 996 | 5,40 | 946 | 0,11 |
| LA17 | | **784** | 812 | 3,57 | **784** | 0,00 |
| LA18 | | **848** | 885 | 4,36 | **848** | 0,00 |
| LA19 | | **842** | 873 | 3,68 | 851 | 1,07 |
| LA20 | | **902** | 912 | 1,11 | 907 | 0,55 |
| LA21 | 15 x 10 | **1046** | 1107 | 5,83 | 1102 | 5,35 |
| LA22 | | **927** | 995 | 7,34 | 974 | 5,07 |
| LA23 | | **1032** | 1049 | 1,65 | 1033 | 0,10 |
| LA24 | | **935** | 1008 | 7,81 | 987 | 5,56 |
| LA25 | | **977** | 1062 | 8,70 | 1028 | 5,22 |
| LA26 | | **1218** | 1296 | 6,40 | 1297 | 6,49 |
| LA27 | | **1235** | 1349 | 9,23 | 1342 | 8,66 |
| LA28 | 20 x 10 | **1216** | 1322 | 8,72 | 1308 | 7,57 |
| LA29 | | **1157** | 1331 | 15,04 | 1286 | 11,15 |
| LA30 | | **1355** | 1410 | 4,06 | 1414 | 4,35 |
| LA31 | | **1784** | **1784** | 0,00 | **1784** | 0,00 |
| LA32 | | **1850** | 1860 | 0,54 | 1884 | 1,84 |
| LA33 | 30 x 10 | **1719** | 1731 | 0,70 | 1723 | 0,23 |
| LA34 | | **1721** | 1778 | 3,31 | 1804 | 4,82 |
| LA35 | | **1888** | 1902 | 0,74 | 1918 | 1,59 |
| LA36 | | **1268** | 1396 | 10,09 | 1352 | 6,62 |
| LA37 | | **1397** | 1517 | 8,59 | 1508 | 7,95 |
| LA38 | 15 x 15 | **1196** | 1315 | 9,95 | 1330 | 11,20 |
| LA39 | | **1233** | 1304 | 5,76 | 1331 | 7,95 |
| LA40 | | **1222** | 1300 | 6,38 | 1338 | 9,49 |

**FIGURA 1.** Comportamiento de los Makespan de los algoritmos para el JSP (Best Known Solution [BKS], Algoritmo Colonia de Hormigas Elitista [EAS], Algoritmo de Selección Clonal [CLONALG])

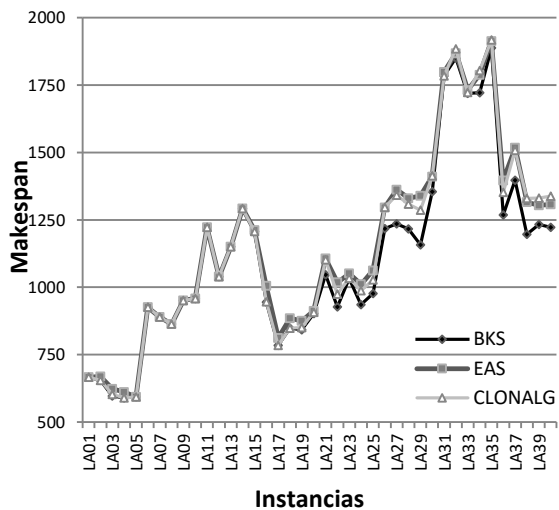

En la figura 2 se observa el comportamiento del error a lo largo de las 40 instancias utilizando los 2 algoritmos, se observa cómo el algoritmo CLONALG presenta el comportamiento más estable de los dos durante las primeras 20 instancias, de ahí en adelante su comportamiento con base al error relativo es similar al algoritmo de colonia de hormigas EAS. Se observa que conforme las instancias son más grandes y de mayor complejidad los errores crecen en la mayoría de los casos haciendo una excepción en la instancia 23 y el rango de la 30 a la 35, este comportamiento es similar en los 2 algoritmos. COMPARARLOS AQUÍ con BKS que no se hizo

**FIGURA 2.** Comportamiento del error relativo para las instancias de Lawrence (Algoritmo Colonia de Hormigas Elitista [EAS], Algoritmo de Selección Clonal [CLONALG])

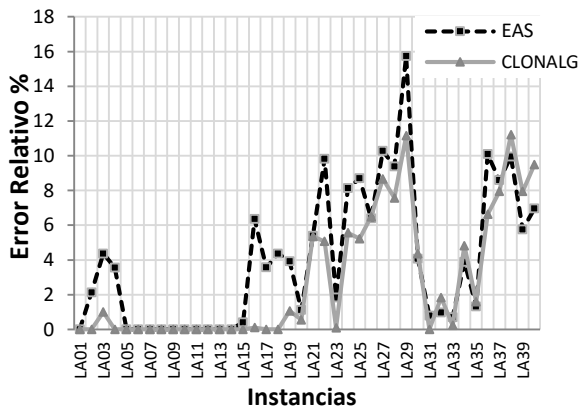

Fuente: Elaboración propia

En la figura 3 se observa la clasificación de los promedios de los errores relativos de cada algoritmo para cada uno de los diferentes tamaños de las instancias de Lawrence, 5 instancias corresponden a cada tamaño y en total son 8 tamaños diferentes por las 40 instancias. Se observa que en los primeros cuatro tamaños se presentan los errores relativos promedios más bajos, siendo el tamaño 15X5 donde los dos algoritmos obtienen el promedio más bajo de los errores relativos y los promedios más altos de los errores relativos se encuentran en las instancias de 20X10 y 15X15. Es interesante observar que aunque en la mayoría de tamaños el mejor rendimiento lo obtuvo el algoritmo CLONALG para el caso de las instancias de mayor tamaño (30x10 y 15X15) presenta un mejor desempeño el algoritmo EAS.

**FIGURA 3.** Representación del error relativo vs tamaño de las instancias para cada algoritmo en el JSP (Algoritmo Colonia de Hormigas Elitista [EAS], Algoritmo de Selección Clonal [CLONALG])

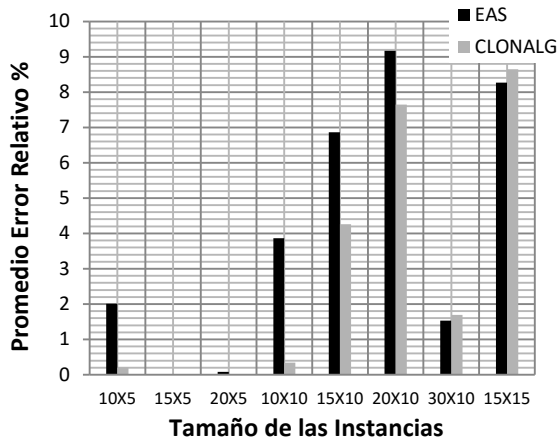

En la Tabla 4 se listan los errores relativos promedios, el mayor error la desviación estándar, y la varianza de cada uno de los algoritmos para el total de las 40 instancias.

**TABLA 4.** Medidas de calidad respecto al error relativo de los algoritmos (Algoritmo Colonia de Hormigas Elitista [EAS], Algoritmo de Selección Clonal [CLONALG])

| Algoritmos | EAS | CLONALG |
|---|---|---|
| **Promedio Error Relativo %** | 3,64 | 2,85 |
| **Mayor Error Relativo %** | 15,04 | 11,20 |
| **Desviación** | 3,90 | 3,66 |
| **Varianza** | 15,24 | 13,41 |

Fuente: Elaboración propia

Se observa que el menor error y la menor dispersión de los datos fueron alcanzados por el algoritmo Inmune artificial, seguido muy cerca por el algoritmo de Hormigas.

En la Tabla 5 se describen la cantidad de mejores soluciones alcanzadas por cada algoritmo y su porcentaje en referencia a las 40 instancias, se observa que el mejor rendimiento fue alcanzado por el algoritmo inmune artificial.

**TABLA 5.** Cantidad de Mejores Soluciones alcanzadas por los algoritmos para el JSP (Algoritmo Colonia de Hormigas Elitista [EAS], Algoritmo de Selección Clonal [CLONALG])

| Algoritmos | EAS | CLONALG |
|---|---|---|
| **Cantidad de BKS alcanzados** | 12 | 17 |
| **Porcentaje BKS alcanzado %** | 30 | 42,5 |

Fuente: Elaboración propia

En la Tabla 6 se comparan los desempeños de los 2 algoritmos respecto a la cantidad de evaluaciones promedio por cada algoritmo y algunas medidas estadísticas descriptivas con base a la dispersión de los datos, se observa que el algoritmo EAS presenta un numero de evaluaciones mucho menor al algoritmo CLONALG y la menor dispersión de datos, se observa que el comportamiento de los algoritmos CLONALG en función de las evaluaciones es muy variable y el consumo de iteraciones de evaluación es alto.

**TABLA 6.** Medidas de desempeño respecto a las evaluaciones de los algoritmos (Algoritmo Colonia de Hormigas Elitista [EAS], Algoritmo de Selección Clonal [CLONALG])

| Algoritmo | EAS | CLONALG |
|---|---|---|
| **Promedio Evaluaciones** | 3124,83 | 54542 |
| **Mayor número de Evaluaciones** | 8016 | 167801 |
| **Menor número de Evaluaciones** | 101 | 4 |
| **Desviación Evaluaciones** | 2557,49 | 47745,49 |
| **Varianza Evaluaciones** | 6540744,866 | 2279631831 |

Fuente: Elaboración propia

## 5. CONCLUSIONES

El presente trabajo realizó una comparación entre los resultados de aplicación a problemas complejos de optimización de dos técnicas bio-inspiradas, que poseen características diferentes en relación a las estrategias de búsqueda que utilizan. Aunque dentro de los procesos de búsqueda de ambas técnicas existe un componente probabilístico, en la primera ese componente es más influyente/marcado durante dicho proceso, en la segunda, una técnica de búsqueda evolutiva, dicho componente permite diversificar la población de soluciones pero la búsqueda se realiza de manera evolutiva.

En la comparación se pudo observar que el algoritmo Inmune Artificial presenta soluciones de mejor calidad que el algoritmo de Colonia de Hormigas para las instancias de Lawrence evaluadas, con un error relativo menor y un porcentaje mayor de mejores soluciones conocidas (BKS) alcanzadas, aunque el algoritmo de Optimización por Colonia de Hormigas presenta un mejor desempeño respecto a la cantidad de evaluaciones requeridas para encontrar la solución y desviación estándar de las mismas , donde la técnica de Optimización por Colonia de Hormigas requiere una menor cantidad de evaluaciones de soluciones que el algoritmo Inmune Artificial, generando soluciones con una búsqueda más eficiente, al converger hacia la solución con menos uso de recursos computaciones y tiempo (esto hace que exijan los tiempos de ejecucion de los algoritmos), lo que la hace adecuada para la exploración de espacios de búsqueda inmensos como los que se presentan en las instancias de mayor tamaño relacionadas a una mayor cercanía a la complejidad problemas reales; *en relación a la calidad de la solución se alcanzaron mejores resultados con la técnica evolutiva.*

Teniendo en cuenta lo anterior, se plantean dos recomendaciones para mejorar el desempeño y la calidad de trabajos futuros, la hibridación de técnicas bio-inspiradas aprovechando los mecanismos de búsqueda más destacadas de cada tipo ( probabilísticas y/o evolutivas), y la adaptación de un modelo de programación en paralelo, aprovechando las funciones e iteraciones que realizan instrucciones con datos diferentes, ejecutándolos sobre la diversidad de arquitecturas de cómputo SMP/SIMD existentes.

## 6. REFERENCIAS BIBLIOGRÁFICAS